\documentclass[10pt,twocolumn,letterpaper]{article}

\usepackage{cvpr}              % To produce the CAMERA-READY version
\usepackage{subcaption}
\usepackage{amsmath}
\usepackage{booktabs}
\usepackage{amssymb}
\usepackage{adjustbox}
\usepackage{makecell}
\usepackage{multirow}
\usepackage{xcolor}
\usepackage{enumitem}% colors
\usepackage{dsfont}
\definecolor{cvprblue}{rgb}{0.21,0.49,0.74}
\usepackage[pagebackref,breaklinks,colorlinks,citecolor=cvprblue]{hyperref}

\def\confName{CVPR}
\def\confYear{2024}

\title{Efficient Test-Time Adaptation of Vision-Language Models}

\author{Adilbek Karmanov\textsuperscript{\rm 1}\thanks{Equal Contribution.} \quad
Dayan Guan\textsuperscript{\rm 2}\footnotemark[1] \quad 
Shijian Lu\textsuperscript{\rm 1,}\textsuperscript{\rm 2}\thanks{Corresponding Author.} \quad  
Abdulmotaleb El Saddik\textsuperscript{\rm 1,}\textsuperscript{\rm 3} \quad  
Eric Xing\textsuperscript{\rm 1,}\textsuperscript{\rm 4} 
\vspace{0.2em} \\
\textsuperscript{\rm 1}Mohamed bin Zayed
University of Artificial Intelligence \quad
\textsuperscript{\rm 2}Nanyang Technological University \\
\textsuperscript{\rm 3}University of Ottawa \quad
\textsuperscript{\rm 4}Carnegie Mellon University \\
}

\begin{document}
\maketitle
\begin{abstract}
Test-time adaptation with pre-trained vision-language models has attracted increasing attention for tackling distribution shifts during the test time. Though prior studies have achieved very promising performance, they involve intensive computation which is severely unaligned with test-time adaptation. We design TDA, a training-free dynamic adapter that enables effective and efficient test-time adaptation with vision-language models. TDA works with a lightweight key-value cache that maintains a dynamic queue with few-shot pseudo labels as values and the corresponding test-sample features as keys. Leveraging the key-value cache, TDA allows adapting to test data gradually via progressive pseudo label refinement which is super-efficient without incurring any backpropagation. In addition, we introduce negative pseudo labeling that alleviates the adverse impact of pseudo label noises by assigning pseudo labels to certain negative classes when the model is uncertain about its pseudo label predictions. Extensive experiments over two benchmarks demonstrate TDA’s superior effectiveness and efficiency as compared with the state-of-the-art. The code has been released in \url{https://kdiaaa.github.io/tda/}.
\end{abstract}

\begin{figure}[!h]
     \centering
      \begin{subfigure}[b]{\linewidth}
      \centering
      \includegraphics[width=\linewidth]{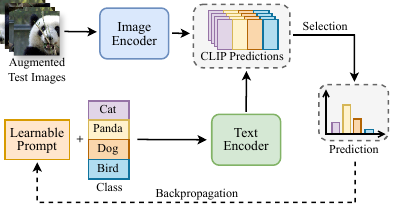}
      \subcaption{Test-time Prompt Tuning~\cite{shu2022testtime,feng2023diverse}}
      \label{fig:tpt}
      \end{subfigure}
  \vfill
  \vspace{5pt}
      \begin{subfigure}[b]{\linewidth}
      \centering
      \includegraphics[width=\linewidth]{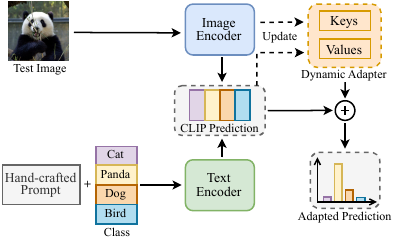}
      \subcaption{Training-free Dynamic Adapter (Ours)}
      \label{fig:short_tda}
      \end{subfigure}
  \caption{
  Comparison of our proposed \textit{Training-free Dynamic Adapter} (TDA) with \textit{Test-time Prompt Tuning} TPT~\cite{shu2022testtime} and its enhancement DiffTPT~\cite{feng2023diverse}: both TPT and DiffTPT require significant computational resources to optimize the \textit{learnable prompt} via backpropagation; TDA is a dynamic cache that is training-free without any backpropagation, making it efficient for test-time adaptation in various real-world scenarios.
  }
  \label{fig_1}
\end{figure}
\section{Introduction}
\label{sec:intro}

Recent advances in vision-language models~\cite{radford2021clip,jia2021scaling,yang2022vision} have opened a new door for integrating human language into various computer vision tasks. Take CLIP~\cite{radford2021clip} as an example. It can enable zero-shot image classification by leveraging a shared embedding space that is learnt from web-scale image-text pairs. Within this shared space, images can be directly recognized by matching their features with the text embeddings of CLIP classes. At the other end, CLIP often faces challenges while handling various specific downstream images, especially when the downstream images have clear domain and distribution shifts as compared with the CLIP training images.

Several recent studies~\cite{wang2021tent,boudiaf2022parameter,chen2022contrastive} introduce a new paradigm called test-time adaptation for mitigating the domain shift. The idea of test-time adaptation is well aligned with real-world scenarios where a model needs to adapt to new environments quickly. Despite its great research and application values, test-time adaptation of vision-language models has been largely neglected. Recently, Test-time Prompt Tuning, as introduced in TPT~\cite{shu2022testtime} and its enhancement DiffTPT~\cite{feng2023diverse}, attempts to adapt vision-language models by learning domain-specific prompts from test data. As illustrated in~\cref{fig_1}~(a), both TPT and DiffTPT train a learnable prompt for each test sample by feeding its augmented views into the CLIP model for generating predictions and minimizing the marginal entropy of confident predictions. Despite its decent performance, the prompt optimization in both TPT~\cite{zhou2022learning} and DiffTPT~\cite{feng2023diverse} is computationally intensive which hinders its applications in various real-world scenarios.

We design a training-free dynamic adapter (TDA) that allows efficient and effective test-time adaptation of vision-language models without requiring any backpropagation in test time. As~\cref{fig_1}~(b) shows, TDA constructs a lightweight \textit{Dynamic Adapter} that keeps a dynamic queue with the pseudo labels of a stream of test samples as values and the corresponding CLIP-extracted features as keys. TDA has two desirable features that make its test-time adaptation highly applicable in real-world scenarios. First, TDA is highly effective, as it improves the quality of pseudo labels via progressive incorporation of test predictions of lower entropy~\cite{wang2021tent, grandvalet2004semi}. Second, TDA is very efficient as the key-value cache is non-parametric and requires no backpropagation during testing. Beyond that, TDA cache is lightweight due to the few-shot setup and it can be computed with simple matrix multiplications ~\cite{grave2017unbounded, Khandelwal2020Generalization,orhan2018simple,zhang2022tip}.

Note that the performance of TDA depends heavily on the pseudo labels of unlabelled test samples which are often noisy with prediction errors. Inspired by the idea of negative learning
~\cite{kim2019nlnl,kim2021joint, rizve2021defense}, we introduce negative pseudo labeling to reduce the impact of noisy estimated labels. Traditional pseudo labeling methods identify the presence of particular classes in unlabeled data, which may result in erroneous pseudo labels being assigned when comparable high probabilities are observed. In contrast, our designed negative pseudo labeling determines the absence of certain classes and can provide more accurate pseudo labels as the probabilities of these complementary classes are very low. Concretely, we construct an additional TDA cache that stores negative pseudo labels to complement the positive TDA cache. By combining positive and negative caches, TDA is more tolerant to noisy pseudo labels and can better generalize to testing data. Extensive experiments over two widely adopted test-time adaptation benchmarks show that TDA outperforms the state-of-the-art by large margins while significantly reducing the testing time from over 12 hours to 16 minutes on the ImageNet dataset.

In summary, the contributions of this work are threefold. \textit{First}, we design a training-free dynamic adapter (TDA) that can achieve test-time adaptation of vision-language models efficiently and effectively. To the best of our knowledge, this is the first work that investigates the efficiency issue of test-time adaptation of vision-language models. \textit{Second}, we introduce negative pseudo labeling to alleviate the adverse impact of pseudo label noises which makes TDA more robust to pseudo label noises and generalizable to testing data. \textit{Third}, we evaluate TDA extensively over two benchmarks, and experiments show that TDA achieves superior accuracy and efficiency compared with the state-of-the-art.

\section{Related Work}
\label{sec:relatedwork}

\noindent\textbf{Vision-language models}~\cite{desai2021virtex, jia2021scaling,  radford2021clip, yang2022vision, hu2022scaling} have demonstrated significant potential in learning semantic representations effectively by undergoing extensive training on image-text data. 
CLIP~\cite{radford2021clip} stands out among these models for its ability to establish links between visual and textual representations, which enables it to achieve impressive zero-shot results on various downstream tasks.
To enhance the transfer learning capability of the CLIP model in the downstream classification tasks, researchers have proposed integrating language prompt learners such as CoOp~\cite{zhou2022learning} and CoCoOp~\cite{zhou2022conditional}, as well as vision adapters such as CLIP-Adapter~\cite{gao2021clip}, Tip-Adapter~\cite{zhang2022tip}, CaFo~\cite{zhang2023prompt}, TaskRes~\cite{yu2023task} and GraphAdapter~\cite{li2024graphadapter}. Although these methods have shown considerable performance improvements, they typically require a large amount of training data in downstream tasks, making them less practical for real-world scenarios. On other hand, this work focuses on a new paradigm named test-time adaptation without accessing the original training data. 

\vspace{10pt}
\noindent\textbf{Test-time adaptation} refers to the process of adapting models to testing data that may have distributional differences from the training data. It is particularly beneficial for real-world applications that require models to be deployed in diverse environments, such as autonomous driving in various weather conditions, medical diagnosis in different hospitals, and etc. Several recent works utilize each batch of testing samples to update partial weights~\cite{sun2020ttt,varsavsky2020test,iwasawa2021test,yi2023temporal}, normalization statistics~\cite{schneider2020improving}, or a combination of both~\cite{wang2021tent,zhang2021adaptive}. 
To avoid updating models with multiple testing samples, MEMO~\cite{zhang2022memo} proposes enforcing the invariant predictions from different augmentations of each sample in the testing data stream.
TPT~\cite{shu2022testtime} tackles the same challenge with vision-language models by fine-tuning a learnable prompt with each testing sample.
DiffTPT~\cite{feng2023diverse} innovates test-time prompt tuning by leveraging pre-trained diffusion models to augment the diversity of test data samples used in TPT.
Although TPT~\cite{shu2022testtime} and DiffTPT~\cite{feng2023diverse} are effective in addressing test-time adaptation of vision-language models, prompt learning is computationally expensive and time-consuming.
This paper aims to mitigate the computational efficiency challenges of TPT and DiffTPT through the introduction of a cache model.

\vspace{10pt}
\noindent\textbf{Cache model} benefits the adaptation techniques by providing efficient inference and non-parametric processing without requiring parameter updates. 
Unbounded Cache~\cite{grave2017unbounded} and PSMM~\cite{merity2016pointer} have shown promising results in the text generation task by storing a large amount of the training dataset to capture long-term dependencies, however, this approach poses a challenge of memory efficiency during the test-time adaptation. A different technique~\cite{johnson2019billion} was proposed to mitigate this limitation by reducing the size of the cache memory through the search find application. Alternatively, Tip-Adapter~\cite{zhang2022tip} solves the cache memory problem by only storing few-shot samples per class to create a cache model for vision-language models. Inspired by this work, our dynamic adapter use the same architecture for the test-time adaptation setting, where there is no access to the source data, and testing samples can only be accessed one by one. To address the lack of access to source data during testing, our adapter collects the most reliable test samples and their pseudo labels to form cache model.

\section{Method}
\label{sec:method}

\subsection{Preliminaries}
\noindent\textbf{CLIP~\cite{radford2021clip}} is a vision-language model that performs a proxy task of predicting the correct pairings of text and image. Consider an $N$-class classification problem where the CLIP's objective in the zero-shot setting is to match images with their most relevant textual descriptions using an image encoder $E_v$ and a text encoder $E_t$. To obtain the textual descriptions, $N$-class names are concatenated with hand-crafted prompts and then mapped into the $c$-channeled text embeddings $\boldsymbol{W}_c$ using the text encoder $E_t$. 

\vspace{10pt}
\noindent\textbf{TPT~\cite{shu2022testtime}} focuses on test-time adaptation of CLIP. 
In TPT, a prompt tuning method is proposed to learn an adaptive prompt $\mathbf{p}_c$ using individual test samples. A set of augmentation functions $\mathcal{A}$ is used to generate $n$ randomly augmented views $\tilde{x}_\mathrm{test} = \mathcal{A}_n(x_\mathrm{test})$ of a test sample $x_\mathrm{test}$.
The objective of TPT is to reduce variation in the model's predictions across different augmentations $\tilde{f}_\mathrm{test} = E_v(\tilde{x}_\mathrm{test})$ by minimizing the marginal entropy among the outputs of the augmented views. 
Furthermore, TPT also includes confidence selection to discard noisy augmentations that could result in ambiguous model predictions, as shown in Figure \ref{fig:tpt}. This is achieved by filtering out augmented views with high-entropy predictions as:
\begin{equation}
    {P_{\rm TPT}(\tilde{f}_\mathrm{test})}
    = \frac{1}{\rho n} \sum_{i=1}^{n}\mathds{1} [\mathrm{H}(\tilde{f_i}\mathbf{p}_c^T) \leq \tau]  \tilde{f_i} \mathbf{p}_c^T,
    \label{eq:tpt}
\end{equation}
where $\mathrm{H}$ is the self-entropy function of the softmax logits predictions, $\tilde{f_i}\mathbf{p}_c^T$ is the class probabilities vector of size $N$ generated from the given $i$-th augmented view of the test image, and the parameter $\tau$ determines that only $\rho$-percentile of confident samples with entropy values below this threshold can be selected out of $n$ augmented views.

\vspace{10pt}
\noindent\textbf{Tip-Adapter~\cite{zhang2022tip}} provides a training-free solution that uses a key-value cache model and integrates knowledge from the pre-trained CLIP model with few-shot labeled samples. The cache model is created using a set of $k$-shot labeled samples $x_k$ from $N$ classes and their corresponding ground-truth labels $y_N$. It can be conceptualized as a linear two-layer model, where the first layer contains train image features $\mathbf{F}_{\rm train} = E_v(x_k)$ and the second layer consists of one-hot vectors $\mathbf{L}_\mathrm{train}$ encoded from the labels $y_N$. 
Given test image features $f_\mathrm{test}$ generated from the CLIP's image encoder $E_{v}$, the prediction from the cache model can be calculated as follows:
\begin{equation}
    % P_{\rm TA}(f_\mathrm{test})
    P_{\rm cache}(f_\mathrm{test})
    = A(f_\mathrm{test} \mathbf{F}^T_{\rm train}) \mathbf{L}_\mathrm{train},
    \label{eq:ta_logits}
\end{equation}
where $A(z) = \alpha \exp(-\beta(1 - z))$ is an adaptation function within a weighting factor $\alpha$ and a sharpness ratio $\beta$. During inference, the prediction of Tip-Adapter is computed by combining the pre-trained CLIP model and the cache model as: $P_{\rm TA}(f_\mathrm{test}) = P_{\rm cache}(f_\mathrm{test}) + f_\mathrm{test} \mathbf{W}_c^T$.

\begin{figure*}[!h]   
  \centering
  \includegraphics[width=\linewidth]{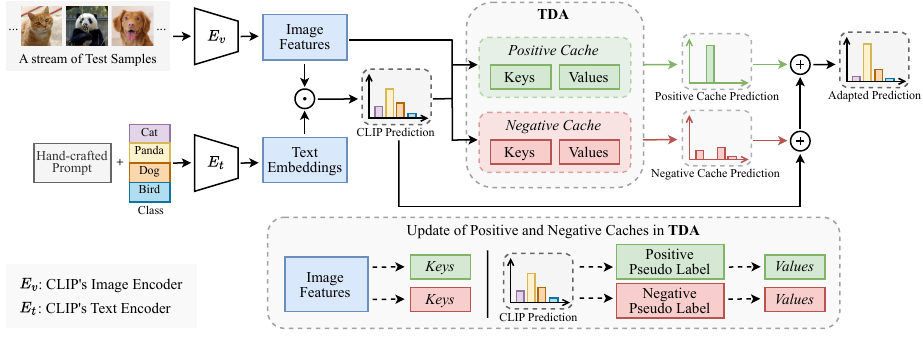}
  \caption{
  Overview of the proposed Training-free Dynamic Adapter (TDA). TDA constructs and updates two key-value caches to store the knowledge of a stream of test samples, and uses the two caches to generate positive and negative predictions which are combined with CLIP predictions to produce the final prediction. Specifically, the CLIP predictions are generated by performing the dot product between the image features generated by CLIP's image encoder $E_v$ and the text embeddings generated by CLIP's text encoder $E_t$, using the 
  hand-crafted prompt and class names. The two key-value caches are updated by gradually incorporating the test features and their corresponding pseudo labels calculated from CLIP's predictions, based on prediction entropy and cache capacity.
  }
  \label{fig_arch}
\end{figure*}

\subsection{Training-free Dynamic Adapter}
During testing, pre-trained vision-language models like CLIP may encounter distribution shifts that degrade the classification performance. To address this issue, existing test-time prompt tuning methods train a learnable prompt by enforcing consistency across different image augmentations during testing. However, it requires a large number of augmentation operations on each test image and computationally-heavy optimization steps to learn the prompt, limiting its applicability in various real-world settings. 

In this paper, our motivation is to design an efficient method for test-time adaptation of the pre-trained vision-language models like CLIP. Inspired by the concept of Tip-Adapter, we propose a training-free dynamic adapter (TDA) to enable efficient and effective test-time adaptation with CLIP. As shown in Figure  \ref{fig_arch}, TDA includes two lightweight key-value caches, where each cache stores a dynamic queue of few-shot test features as keys and the corresponding pseudo labels as values. The first cache is intended for positive learning and it dynamically updates key-value pairs with high-confidence predictions to improve the accuracy. The second cache is designed for negative learning and it aims to address the adverse effects of noisy pseudo labels by introducing negative pseudo labeling to identify class absence rather than presence.
By combining the positive cache and negative cache, the proposed TDA can achieve superior performance in terms of both speed and accuracy.

Our goal is to conduct test-time adaptation by gathering adequate knowledge from the testing data stream and improving the predictions through the adaptation of the image features. To accomplish this goal, we create positive and negative caches that capture particular characteristics of the testing data stream. In the upcoming parts, we will present the process of collecting data for each cache and define the conditions in which we can utilize cache information to adapt features and enhance model predictions.

\paragraph{Positive Cache.} The positive cache in TDA is a key-value cache, in which keys and values are represented as a dynamic queue. It aims to collect high-quality few-shot pseudo labels $\mathbf{\hat{L}_p}$ as positive values and the corresponding features $\mathbf{Q_{p}}$ as keys. The key-value cache is initially empty and then accumulates a sufficient number of key-value pairs during the test-time adaptation. To maintain high-quality pseudo labels, TDA progressively incorporates test predictions with lower entropy while limiting shot capacity\footnote{The shot capacity refers to the maximum number of pairs per class.} in the positive cache. Different from a normal queue like a FIFO queue with a fixed size, the dynamic queue in our method expands in size during testing. Besides, the dynamic queue operates similarly to a priority queue, using entropy as the criterion for prioritization. Note that, each class has its own queue to maintain the order and correct data structure of each class in the cache.

Given a pre-trained CLIP model that consists of a text encoder $E_t$ and an image encoder $E_v$, $E_t$ processes class names with pre-defined prompts to generate $c$-channeled text embeddings $\mathbf{W}_c$ and $E_v$ processes each test image $x_\mathrm{test}$ to produce image features $f_\mathrm{test}$.
To build the positive cache, TDA first generates a pseudo label $\hat{l}$, a one-hot encoded vector of a categorical distribution, for each test sample $x_{\rm test}$ by applying the softmax function to the prediction $f_{\rm test} \mathbf{W}_{c}^{T}$. We establish two conditions to ascertain whether and how to include the pseudo label $\hat{l}$ and its corresponding image features $f_{\rm test}$ into the positive cache. The first condition is defined as: if the shot number (the number of collected pairs per class) of $\mathbf{\hat{L}_p}$ is less than the maximum shot capacity (the maximum capacity number of pairs per class) $k$, TDA will add $\hat{l}$ and $f_{\rm test}$ as a new value and a new key to $\mathbf{\hat{L}_p}$ and $\mathbf{Q_{p}}$ respectively. Meanwhile, the second condition is defined as: if the shot number of $\mathbf{\hat{L}_p}$ has reached the maximum shot capacity $k$, TDA will replace an `uncertain' key-value pair $\{\mathbf{q}^{ent},\hat{l}_{p}^{ent}\}$ with $\{f_{\rm test}, \hat{l}\}$ when $\mathrm{H}(f_{\rm test} \mathbf{W}_c^T) < \mathrm{H}(\mathbf{q}^{ent} \mathbf{W}_c^T)$. Here, $\mathrm{H}$ denotes the entropy function and the term `uncertain'  indicates that the entropy of a particular key-value pair is the highest compared to the entropy of all other key-value pairs of the same class in the cache model. By applying these two conditions, TDA can gradually integrate test predictions with lower entropy while controlling the shot capacity, which helps to ensure the collection of high-quality pseudo labels in positive cache. 

During test-time adaptation, the positive cache can quickly retrieve relevant information by treating the image features $f_\mathrm{test}$ generated from a test example as a query and searching the stored key-value pairs $\{\mathbf{Q_{p}},\mathbf{\hat{L}_p}\}$ for matching information. The adapted predictions using the positive cache can then be obtained as:
\begin{equation}
    P_{\rm pos}(f_\mathrm{test}) =  A (f_\mathrm{test} \mathbf{Q_{p}^T}) \mathbf{\hat{L}_p},
    \label{eq:pos_tda}
\end{equation}
where $A$ is the adaptation function defined in Tip-Adapter.

\paragraph{Negative Cache.} Similar to the positive cache in our TDA, the negative cache is also a dynamic queue structure with negative keys and negative values denoted as $\mathbf{Q_{n}}$ and $\mathbf{\hat{L}_n}$, respectively. It aims to gather CLIP-generated image features to $\mathbf{Q_{n}}$ and the corresponding negative pseudo labels to $\mathbf{\hat{L}_n}$. Unlike the pseudo labels in the positive cache, the negative pseudo labels are obtained by applying negative mask on the class probabilities as:
\begin{equation}
    \mathbf{\hat{L}_n} = - \mathds{1}[p_l < P(\mathbf{Q_{n}})],
    \label{eq:neg_mask}
\end{equation}
where higher probabilities than $p_l$ are selected as negative pseudo labels from uncertain predictions and the uncertainty is measured by the entropy of predictions. 
Here, $p_{l}$ represents a threshold in negative pseudo labeling, and $\mathbf{\hat{L}_n}$ denotes a negative pseudo label, which is a vector whose elements larger than $p_l$ have a value -1 and otherwise 0.
Different from existing negative learning methods~\cite{kim2019nlnl,kim2021joint, rizve2021defense} that select negative labels from all noisy labels, TDA selects negative pseudo labels from uncertain predictions to avoid bias to the data with certain predictions.

When constructing the negative cache, the testing feature $f_\mathrm{test}$ will 
be included in negative cache if it satisfies the condition $\gamma(f_\mathrm{test})$: the entropy of the prediction is in the specified interval between $\tau_l$ and $\tau_h$:
\begin{equation}
\gamma(f_\mathrm{test}): \tau_l < \mathrm{H}(f_\mathrm{test} \mathbf{W}_c^T) < \tau_h.
\label{eq:unconfidence}
\end{equation}
This condition is designed to mitigate the risk of prediction errors due to high entropy or be biased to certain predictions (characterized by very low entropy), by incorporating test samples that exhibit a moderate degree of prediction uncertainty. Once the $\gamma(f_\mathrm{test})$ check is completed, the remaining steps for collecting uncertain samples in the negative cache follow the same two conditions designed in the positive cache. Similar to the positive cache, TDA also limits the shot capacity $\tilde{k}$ in the negative cache. 

During test-time adaption, the testing features $f_\mathrm{test}$ can be quickly adapted to target domains by retrieving the knowledge from $\{\mathbf{Q_{n}},\mathbf{\hat{L}_n}\}$ in the negative cache and the adapted prediction can be obtained as:
\begin{equation}
P_{\rm neg}(f_\mathrm{test}) = -A (f_\mathrm{test} \mathbf{Q_{n}^T}) \mathbf{\hat{L}_n},
\label{eq:neg_tda}
\end{equation}
where $A$ is the adaptation function defined in Tip-Adapter.
The predictions of TDA can be formulated by combining the negative cache, the positive cache and the pre-trained CLIP model together as follows:
\begin{equation}
   P_\mathrm{TDA}(f_\mathrm{test}) = f_\mathrm{test} \mathbf{W}_c^T + P_{\rm pos} (f_\mathrm{test}) +  P_{\rm neg}(f_\mathrm{test}).
    \label{eq:final}
\end{equation}

\renewcommand\arraystretch{0.9}
\begin{table*}[ht]
  \centering
  \resizebox{\linewidth}{!}{
  \begin{tabular}{l*{7}c}
    \toprule
    {Method}      & ImageNet   &  ImageNet-A  &  ImageNet-V2  & ImageNet-R 
    & ImageNet-S   
    &{Average}  & {OOD Average}     \\
  \midrule
  CLIP-ResNet-50       &59.81&	23.24&	52.91	&{60.72}	&35.48&	46.43&	43.09           \\ 
  \midrule
  CoOp            &  \textbf{63.33} &  23.06  &   55.40     &  56.60   &  34.67   &     46.61   &     42.43       \\
  
  CoCoOp         &  62.81 &  23.32  &   55.72     &  57.74  &  34.48   &     46.81   &    42.82   \\
Tip-Adapter & 62.03 & 23.13 & 53.97 & 60.35  & 35.74 & 47.04 & 43.30 \\
  \midrule
  TPT          &  60.74 & 26.67  &    54.70   &  59.11   &  35.09   &    47.26    &     43.89    \\
  DiffTPT & 60.80 & \textbf{31.06} & \textbf{55.80} & 58.80 & 37.10 & 48.71 & 45.69 \\
    \textbf{TDA (Ours)} & 61.35 & {30.29} & 55.54
    & \textbf{62.58} & \textbf{38.12} & \textbf{49.58} & \textbf{46.63} \\
  \midrule
  \midrule
  CLIP-ViT-B/16       & 68.34&	49.89&	61.88&	{77.65}&	48.24&	61.20&	59.42             \\
  \midrule
  CoOp          &  \textbf{71.51} &  49.71  &   64.20     &  75.21   &  47.99   &   61.72     &   59.28  \\
  
  CoCoOp         &  71.02  & 50.63  & 64.07       &  76.18   &  48.75   &   62.13     &   59.91  \\
 Tip-Adapter& 70.75 & 51.04 & 63.41 & 77.76 & 48.88 & 62.37 & 60.27 \\
  \midrule
  TPT           &  68.98  & 54.77  & 63.45       &  77.06   &  47.94   &   62.44     &   60.81 \\
  DiffTPT           &  70.30  & 55.68  & \textbf{65.10}       &  75.00   &  46.80   &   62.28     &   60.52 \\
  \textbf{TDA (Ours)} & 69.51 & \textbf{60.11} & 64.67 & \textbf{80.24} & \textbf{50.54} & \textbf{65.01} & \textbf{63.89} \\
    \bottomrule
  \end{tabular}
}
\caption{\textbf{Results on the OOD Benchmark}. Our TDA is compared with several state-of-the-art methods designed for vision-language models: the baseline method CLIP, three train-time adaptation methods (\textit{i.e.}, CoOp, CoCoOp, and Tip-Adapter), and two test-time adaptation methods (\textit{i.e.}, TPT and DiffTPT). All the compared methods are built upon CLIP-ResNet-50 or CLIP-ViT-B/16 baselines. The two evaluation metrics \textit{Average} and \textit{OOD Average} are calculated by taking the mean accuracy across all five datasets and four OOD datasets excluding ImageNet. The results of CLIP, CoOp, CoCoOp, and TPT are obtained from the TPT paper, the results of DiffTPT are obtained from the DiffTPT paper,
while the results of Tip-Adapter are reproduced using the official codes.
}
\label{tab:ood-main}
\end{table*}
  
\renewcommand\arraystretch{1.0}
\begin{table}[h]
\centering
\begin{tabular}{lccccc}
\toprule
    Method & Testing Time &  Accuracy &Gain\\ \midrule
    CLIP-ResNet-50 &\underline{\textbf{12min}} &59.81 &0\\
    TPT &12h\ 50min & 60.74  &{+0.93}\\
    DiffTPT &34h\ 45min & 60.80  &{+0.99}\\
    \textbf{TDA (Ours)} & \textbf{16min} & \textbf{61.35}\ &\textbf{+1.54}\\
\bottomrule
\end{tabular}
\caption{Comparisons of our TDA with CLIP-ResNet-50, TPT, and DiffTPT in terms of efficiency (\textit{Testing Time}) and effectiveness (\textit{Accuracy}). 
The final column shows the accuracy gain relative to the baseline CLIP. Note that the testing time of DiffTPT does not include the duration required for the image generation process with pre-trained diffusion models, which is an additional time-consuming factor during the testing phase.
}
\label{fig:long}
\end{table}

\subsection{Relationship with TPT and Tip-Adapter}

Both  TPT and TDA are designed to handle the challenge of adapting models to test data that have distributional discrepancies from the training data. 
TPT trains a learnable prompt $\mathbf{p}_c$ in~\cref{eq:tpt} for each test sample with a large number of augmentations and such training process requires backpropagation and is computationally intensive. In contrast, our TDA is training-free as both positive cache \{$\mathbf{Q_{p}}$, $\mathbf{\hat{L}_p}$\} and negative cache \{$\mathbf{Q_{n}}$, $\mathbf{\hat{L}_n}$\} are non-parametric, which is super-efficient without incurring any backpropagation. 

Our TDA employs a cache model that shares similarities with the Tip-Adapter, where features and labels are stored as key-value pairs in a memory cache. 
One notable distinction between the Tip-Adapter and TDA lies in the type of cache used. Specifically, the Tip-Adapter relies on a static cache \{$\mathbf{F}_{\rm train}, \mathbf{L}_{\rm train}$\} because it is designed for a supervised adaptation setting, where the ground-truth labels $\mathbf{L}_{\rm train}$ are predetermined and readily available. In contrast, our TDA introduces a new dynamic cache \{$\mathbf{Q_{p}}$, $\mathbf{\hat{L}_p}$\} designed for a test-time adaptation setting, where the pseudo labels $\mathbf{\hat{L}_p}$ are generated on-the-fly from a stream of test samples. Furthermore, TDA incorporates a novel negative cache model $\{\mathbf{Q_{n}},\mathbf{\hat{L}_n}\}$ that enhances testing predictions by utilizing indirect knowledge that a test image does not belong to certain negative classes. By combining positive and negative caches, our proposed TDA is more robust to noisy pseudo labels and can generalize well to testing data.

%%%%%%%%%%%%%%%%%%%%%%%%%%

\renewcommand\arraystretch{0.95}
\begin{table*}[ht]
  \centering
  \resizebox{\linewidth}{!}{
    \begin{tabular}{l*{11}c}
      \toprule
      Method & Aircraft & Caltech101 & Cars & DTD & EuroSAT & Flower102 & Food101 & Pets & SUN397 & UCF101 & Average \\
      \midrule
      CLIP-ResNet-50 & 16.11 & 87.26 & 55.89 & 40.37 & 25.79 & 62.77 & 74.82 & 82.97 & 60.85 & 59.48 & 56.63 \\
      \midrule
      CoOp & 15.12 & 86.53 & 55.32 & 37.29 & 26.20 & 61.55 & 75.59 & 87.00 & 58.15 & 59.05 & 56.18 \\
      CoCoOp & 14.61 & 87.38 & 56.22 & 38.53 & 28.73 & 65.57 & 76.20 & \textbf{88.39} & 59.61 & 57.10 & 57.23 \\
      \midrule
      TPT & 17.58 & 87.02 & 58.46 & 40.84 & 28.33 & 62.69 & 74.88 & 84.49 & 61.46 & 60.82 & 57.66 \\
      DiffTPT & 17.60 & 86.89 & \textbf{60.71} & 40.72 & 41.04 & 63.53 & \textbf{79.21} & 83.40 & \textbf{62.72} & 62.67 & 59.85 \\
       \textbf{TDA (Ours)}  & \textbf{17.61} & \textbf{89.70} & 57.78 & \textbf{43.74} & \textbf{42.11} & \textbf{68.74} & 77.75 & 86.18 & 62.53 & \textbf{64.18}  & \textbf{61.03}\\
      \midrule
      \midrule
      CLIP-ViT-B/16 & 23.22 & 93.55 & 66.11 & 45.04 & 50.42 & 66.99 & 82.86 & 86.92 & 65.63 & 65.16 & 64.59 \\
      \midrule
      CoOp & 18.47 & 93.70 & 64.51 & 41.92 & 46.39 & 68.71 & 85.30 & 89.14 & 64.15 & 66.55 & 63.88 \\
      CoCoOp & 22.29 & 93.79 & 64.90 & 45.45 & 39.23 & 70.85 & 83.97 & \textbf{90.46} & 66.89 & 68.44 & 64.63 \\
      \midrule
      TPT & 24.78 & 94.16 & 66.87 & \textbf{47.75} & 42.44 & 68.98 & 84.67 & 87.79 & 65.50 & 68.04 & 65.10 \\
      DiffTPT & \textbf{25.60} & 92.49 & 67.01 & 47.00 & 43.13 & 70.10 & \textbf{87.23} & 88.22 & 65.74 & 62.67 &65.47 \\
      \textbf{TDA (Ours)} & 23.91 & \textbf{94.24} & \textbf{67.28} & 47.40 & \textbf{58.00} & \textbf{71.42} & 86.14 & 88.63 & \textbf{67.62} & \textbf{70.66} & \textbf{67.53} \\
      \bottomrule
    \end{tabular}
  } 
  \caption{\textbf{Results on the Cross-Domain Benchmark.} Our TDA is compared with several state-of-the-art methods designed for vision-language models: the baseline method CLIP, two train-time adaptation methods (\textit{i.e.}, CoOp and CoCoOp), and two test-time adaptation methods (\textit{i.e.}, TPT and DiffTPT). Note that Tip-Adapter is unable to be evaluated on the Cross-Domain Benchmark as it cannot handle new classes during testing. The evaluation metric \textit{Average} is calculated by taking the mean accuracy across all ten datasets. 
  The results of CLIP, CoOp, CoCoOp, and TPT are obtained from the TPT paper, while the results of DiffTPT are obtained from the DiffTPT paper.
  }
  \label{tab:fine-grained}
  \vspace{10pt}
\end{table*}

%%%%%%%%%%%%%%%%%%%%%%%%%%%
  
\section{Experiments}
\subsection{Experimental Setup}
\paragraph{Benchmarks.} 
We conducted main experiments on two benchmarks: out-of-distribution (OOD) benchmark and cross-domain benchmark, both applied in the previous work~\cite{shu2022testtime} that adapts vision-language models in test time.
The OOD benchmark serves as a measure of the robustness of our approach by involving assessment on 4 out-of-distribution datasets derived from ImageNet~\cite{imagenet_cvpr09}: ImageNet-A~\cite{Hendrycks2021ima}, ImageNet-V2~\cite{Recht2019imv2}, ImageNet-R~\cite{Hendrycks2021imr}, and ImageNet-S~\cite{wang2019learning}. This benchmark is specifically designed to evaluate a model's capacity to generalize to new and unseen data. 
The cross-domain benchmark, on the other hand, is involved to evaluate the model's performance across 10 diverse image classification datasets, each from a distinct domain with different classes: Aircraft~\cite{aircrafts}, Caltech101~\cite{caltech101}, Cars~\cite{stanfordcars}, DTD~\cite{dtd}, EuroSAT~\cite{eurosat}, Flower102~\cite{flowers102}, Food101~\cite{food101}, Pets~\cite{oxfordpets}, SUN397~\cite{sun397}, and UCF101~\cite{ucf101}. This benchmark provides a comprehensive evaluation of the model's adaptability during test time across various class spaces.

\paragraph{Implementation details.}
\label{sec:impl_details}
All the models in our experiments are built upon the pre-trained CLIP model~\cite{radford2021clip} that consists of an image encoder and a text encoder. The image encoder can be either ResNet~\cite{resnet} or Vision Transformer~\cite{dosovitskiy2020image}, while the text encoder is Transformer~\cite{vaswani2017attention}.
Test-time adaptation is set for single-image scenarios, using a batch size of 1.
We conduct a search for all our hyperparameters using a single ImageNet validation set. 
The threshold $p_{l}$ for negative pseudo-labeling in Eq~4 is set as 0.03. The upper and lower thresholds [$\tau_l$, $\tau_h$] for testing feature selection in Eq~5 are set as [0.2, 0.5].
Once searched, these hyperparameters are fixed and evaluated across various new datasets. To avoid incurring backpropagation when using learnable prompt, we follow~\cite{radford2021clip} to use hand-crafted prompts. 
We use top-1 accuracy (\%), a standard classification criterion, as our evaluation metric.
All the experiments are conducted using a single NVIDIA Quadro RTX 6000 GPU.

\subsection{Comparisons with State-of-the-art}
In this section, we compare our proposed TDA with several state-of-the-art methods, including CLIP \cite{radford2021clip}, three train-time adaptation methods, \textit{i.e.}, CoOp \cite{zhou2022learning}, CoCoOp \cite{zhou2022conditional}, and Tip-Adapter \cite{zhang2022tip}, as well as 
two existing test-time adaptation methods TPT \cite{shu2022testtime} and DiffTPT~\cite{feng2023diverse}, 
all of which are designed for vision-language models. Specifically, CLIP is evaluated using an ensemble of 80 hand-crafted prompts as in \cite{radford2021clip}. All train-time adaptation methods are trained on the ImageNet train set with 16 shots per class and tested on other datasets as in \cite{shu2022testtime}. CoOp utilizes a learnable context module of size 4 that is fine-tuned for specific downstream tasks, while CoCoOp is the generalized version of CoOp with the input-conditional token for image features. Tip-Adapter employs the optimal hyperparameters obtained from the ImageNet validation set during evaluation. We would like to note that Tip-Adapter is unable to handle new classes during testing, limiting its implementation to OOD benchmark evaluation where the training classes encompass all the testing classes. 
Different from train-time adaptation methods, test-time adaptation methods (\ie, TPT, DiffTPT, and our TDA) do not utilize the ImageNet train set. Instead, they are fine-tuned with target datasets using a stream of unlabeled test samples. Following TPT and DiffTPT, we compare TDA with the state-of-the-art over two public benchmarks: OOD benchmark and cross-domain benchmark.

\paragraph{Results on the OOD Benchmark.}
We first compare TDA with state-of-the-art methods over the OOD benchmark. Table \ref{tab:ood-main} presents the experimental results, highlighting the superior performance of the proposed TDA compared to both TPT and DiffTPT across various OOD datasets derived from ImageNet. Specifically, TDA outperforms TPT on both ResNet-50 and ViT-B/16 architectures, improving OOD accuracy by 2.74\% and 3.08\% on average, respectively. Furthermore, compared to DiffTPT, TDA exhibits an average accuracy improvement of 0.94\% and 3.37\% in the OOD benchmark for ResNet-50 and ViT-B/16, respectively. 
These results validate the effectiveness of TDA in enhancing test-time adaptation performance on various OOD test datasets.

In order to provide a more comprehensive evaluation of our proposed method's efficiency and effectiveness, we compared it with the baseline CLIP-ResNet-50 and two existing test-time adaptation methods (\ie, TPT and DiffTPT).
This comparison encompasses both testing time and testing accuracy, and the corresponding results are shown in Table \ref{fig:long}. 
This evaluation is performed on the ImageNet validation dataset, which consists of 50,000 images, using a single NVIDIA Quadro RTX 6000 GPU. 
When compared to the baseline CLIP-ResNet-50, the proposed TDA demonstrates a significant improvement in testing accuracy (+1.54\%), with only a minimal sacrifice in testing efficiency (requiring an additional 4min). 
In comparison to TPT and DiffTPT, the proposed TDA demonstrates not only superior testing accuracy but also significantly improved efficiency. It reduces the testing time dramatically from 12h~50min by TPT and even more from 34h~45min by DiffTPT, down to just 16 minutes. 
Without including the image generation time, DiffTPT consumes clearly more test time than TPT as it involves a time-consuming multi-step prompt updating process whereas TPT requires a single step only.
The experimental results strongly validate the effectiveness and efficiency of our proposed method, establishing its suitability for real-world applications.

We then compare TDA with state-of-the-art methods over cross-domain benchmark. The results, presented in Table~\ref{tab:fine-grained}, demonstrate that TDA not only surpasses the performance of the TPT method but also shows a significant advantage over its improvement method DiffTPT.
Specifically, when utilizing CLIP-ResNet-50 and CLIP-ViT-B/16 as the backbone, TDA achieves an improvement in average accuracy over TPT by 3.37\% and 2.43\%, respectively. These improvements, along with a 1.18\% and 2.06\% gain over DiffTPT for the respective backbones, further verify the effectiveness of TDA in adapting to diverse class datasets during test time. This attribute holds significant value for vision-language models such as CLIP, as it enables them to classify arbitrary classes in image classification without the need for additional training.

\begin{figure}[!t]
\centering
\includegraphics[width=\linewidth]{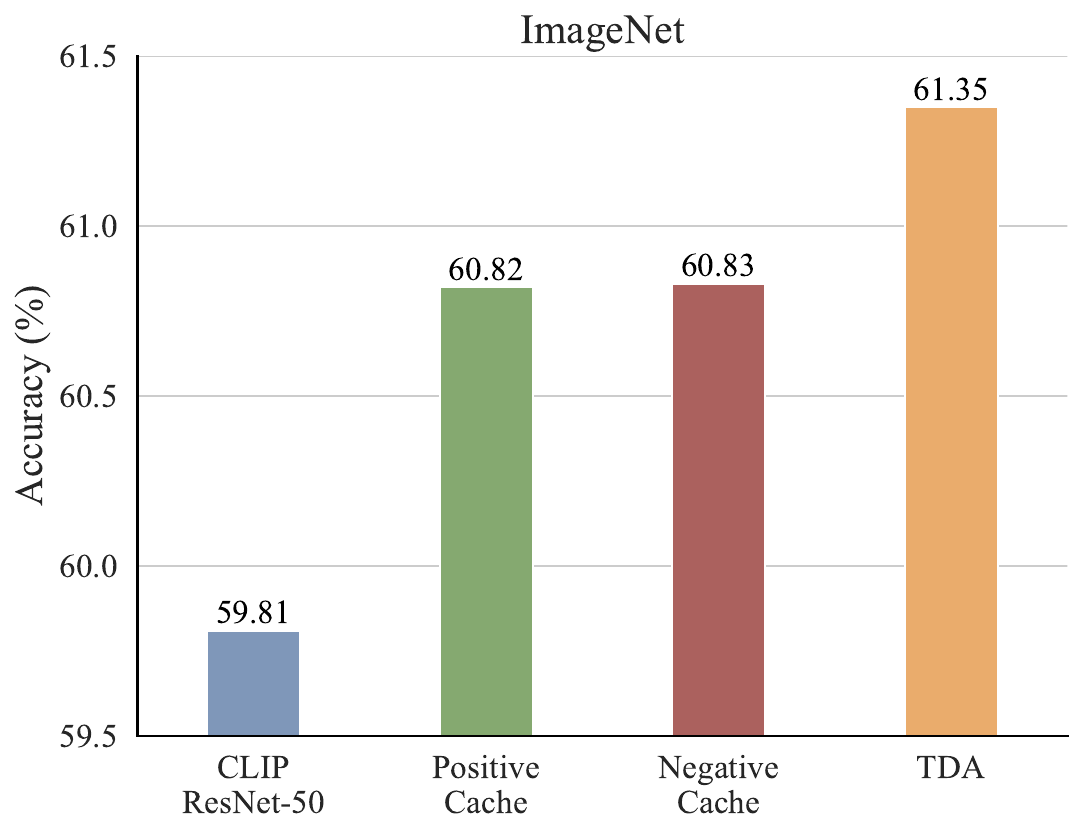}
  \vfill
  \vspace{-5pt}
  \caption{Ablation studies on two cache designs in TDA: \textit{Positive Cache} and \textit{Negative Cache}. All the models are built upon the baseline model CLIP-ResNet-50.}
\label{fig:combination}
\end{figure}

\subsection{Ablation Studies}
\label{par:ablation}

In this section, we perform ablation studies to examine the effectiveness of our designs. All the ablation studies are conducted over the ImageNet dataset, where TDA can achieve an accuracy of 61.35\% under default settings. TDA consists of a \textit{Positive Cache} and a \textit{Negative Cache}, which perform positive and negative pseudo labeling within our designed dynamic adapter, respectively. We first assess the efficacy of the two cache designs in TDA. As shown in Figure \ref{fig:combination}, both \textit{Positive Cache} and \textit{Negative Cache} significantly surpass the baseline model CLIP, demonstrating that test-time adaptation can be improved by introducing a dynamic adapter with either positive pseudo labeling or negative pseudo labeling. Besides, the two cache designs in TDA can complement each other as TDA (\textit{i.e.}, the combination of the two designs) clearly outperforms either \textit{Positive Cache} or \textit{Negative Cache} on the challenging ImageNet dataset. 
Moreover, we extend our ablation studies to the cross-domain benchmark, and results show that the positive cache achieved 60.38\% accuracy, the negative cache 60.11\%, while their combination yielded a 61.03\% accuracy, thereby highlighting the significance of each type of cache in enhancing TDA's performance.

\begin{figure}[!t]
\centering
\includegraphics[width=\linewidth]{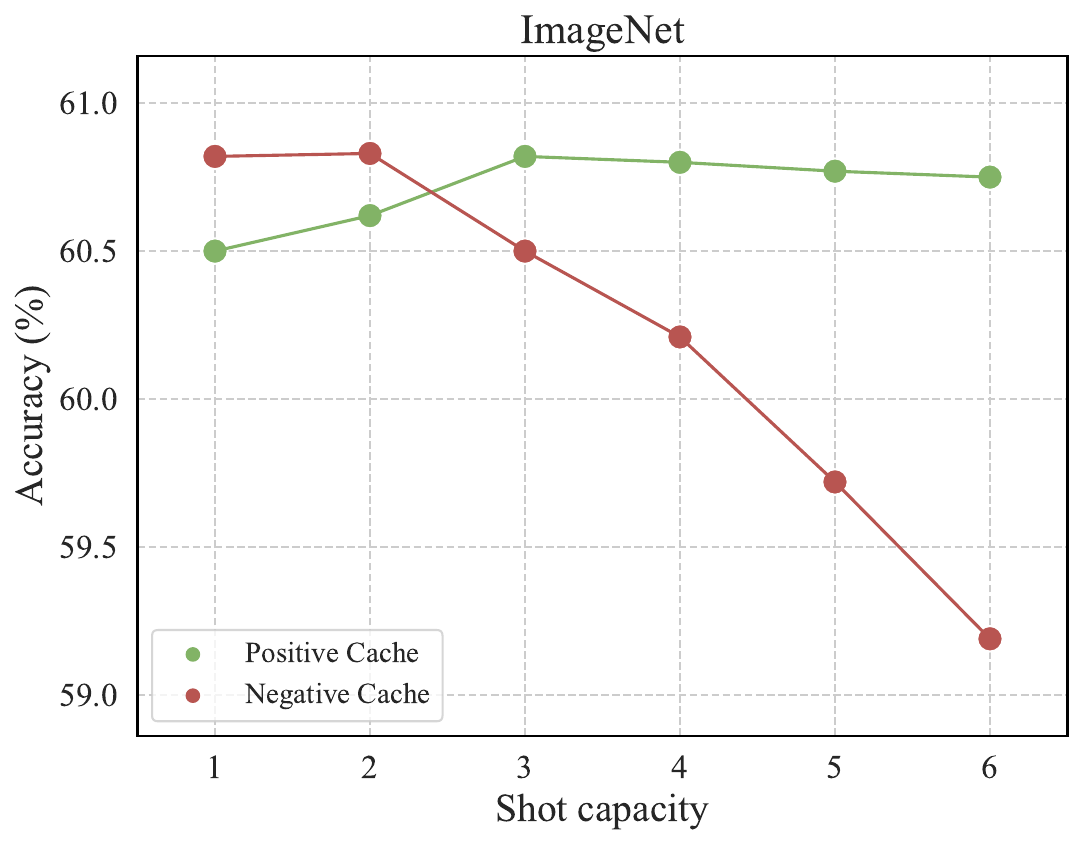}
  \vfill
    \vspace{-5pt}
  \caption{Parameter studies on the \textit{Shot Capacity} in \textit{Positive Cache} and \textit{Negative Cache}. 
  }
\label{fig:capacity}
\end{figure}

We proceed to perform parameter studies on the shot capacity, which refers to the maximum number of key-value pairs per class, in both \textit{Positive Cache} and \textit{Negative Cache} models. These studies aims to find the optimal balance between the diversity and accuracy of the key-value pairs. Figure \ref{fig:capacity} shows that the performance of both cache models is significantly affected when the shot capacity is either too low or too high. We find that the shot capacity is set as 3 for the \textit{Positive Cache} and 2 for the \textit{Negative Cache} yields the best performance. This is because an appropriate shot capacity ensures both high-quality pseudo labels (paired values) and diverse image features (paired keys) in the cache models. 
The negative cache is sensitive to shot numbers due to its role in storing probabilities for various negative pseudo labels, each representing a class to be excluded from the model. 
Contrary to the intuition that a larger negative cache might be beneficial, a larger negative cache leads to more high-entropy, noisier pseudo labels, as highlighted in self-training studies like~\cite{grandvalet2004semi}, thereby lowering the performance.
Conversely, the positive cache, which stores single and high-confidence prediction, is less affected by variations in shot capacity, thereby maintaining consistent accuracy across different shot capacities.
To facilitate practical applications, the shot capacity settings are fixed and directly applied to new datasets without the need for additional parameter adjustments.

\section{Conclusion}
In this work, we have presented TDA, a dynamic adapter for efficient and effective test-time adaptation of vision-language models. The proposed method employs a key-value cache, which maintains a dynamic queue with test-sample features as keys and corresponding few-shot pseudo labels as values, allowing for gradual adaptation to test data through progressive pseudo label improvement. Moreover, TDA introduces a negative cache to mitigate the undesirable effects of noisy pseudo labels by assigning negative pseudo labels to certain classes when the model is uncertain about its predictions. The results of extensive experiments over two benchmarks demonstrate that TDA outperforms state-of-the-art test-time adaptation methods while significantly reducing testing time. This work contributes to the research and application values of test-time adaptation and presents a promising solution to the efficiency issue of test-time adaptation of vision-language models.

\section*{Acknowledgement}
This study is supported under the Mohamed bin Zayed University of Artificial Intelligence.

{
    \small
    \bibliographystyle{ieeenat_fullname}
    \bibliography{main}
}

\newpage

\section{Benchmark Details}
This section provides detailed information on the two benchmarks used in our work. 
\vspace{6pt}
\noindent \textbf{OOD Benchmark} 
is used to evaluate the model robustness against natural distribution shifts using the traditional ImageNet and its out-of-distribution (OOD) versions containing images with varying styles and corruptions. Herein below, we provide a concise overview of each of the OOD datasets.
\begin{itemize}[leftmargin=*]\setlength\itemsep{-0em}
    \item \textbf{ImageNet-V2}~\cite{Recht2019imv2} consists of 10,000 images and 1,000 ImageNet classes, and was collected by applying an updated natural data collection pipeline to the original ImageNet data.
    \item \textbf{ImageNet-A}~\cite{Hendrycks2021ima}  is a subset of 7,500 visually similar but naturally perturbed ImageNet images of 200 classes.
    \item \textbf{ImageNet-R}~\cite{Hendrycks2021imr} includes 30,000 images belonging to 200 categories of the ImageNet dataset, but with diverse artistic styles.
    \item \textbf{ImageNet-S}~\cite{wang2019learning} 
    consists of 50,000 sketches of 1000 class objects from the ImageNet dataset, and represents a domain shift from natural images to sketches.
\end{itemize}

\vspace{6pt}
\noindent \textbf{Cross-Domain Benchmark} 
consists of 10 image classification datasets to evaluate the effectiveness of the method on different domains. This benchmark incorporates the following datasets: Caltech101 \cite{caltech101} for general image classification, OxfordPets (Pets) \cite{oxfordpets}, StanfordCars (Cars) \cite{stanfordcars}, Flowers102 \cite{flowers102}, Food101 \cite{food101}, and FGVCAircraft (Aircraft) \cite{aircrafts} for fine-grained image classification, EuroSAT \cite{eurosat} for satellite image classification, UCF101 \cite{ucf101} for action classification, DTD \cite{dtd} for texture classification, and SUN397 \cite{sun397} for scene classification.

\noindent The detailed statistics of all the datasets are shown in Table~\ref{tab:datasets}.

\begin{table}[h]
    \centering
    \renewcommand\arraystretch{1.1}{
    \begin{tabular}{l c c}
    \toprule
    Dataset & Classes & Test Size\\
    \midrule
    ImageNet & 1,000 & 50,000 \\
    ImageNet-V2 & 1,000  & 10,000\\
    ImageNet-S  & 1,000 & 
    50,000\\
    ImageNet-A & 200  & 7,500 \\
    ImageNet-R & 200 & 30,000 \\
    \midrule
    Aircraft & 100 & 3,333 \\
    Caltech101 & 100 & 2,465 \\
    Cars & 196 & 8,041 \\
    DTD & 47 & 1,692 \\
    EuroSAT & 10 & 8,100 \\
    Flowers102 & 102 & 2,463 \\
    Food101 & 101 & 30,300 \\
    Pets & 37 & 3,669 \\
    SUN397 & 397 & 19,850 \\
    UCF101 & 101 & 3,783 \\
    \bottomrule
    \end{tabular}}
    \caption{Datasets statistics.}
    \label{tab:datasets}
\end{table}

\begin{figure}[h]
  \centering
  \includegraphics[scale=0.46]{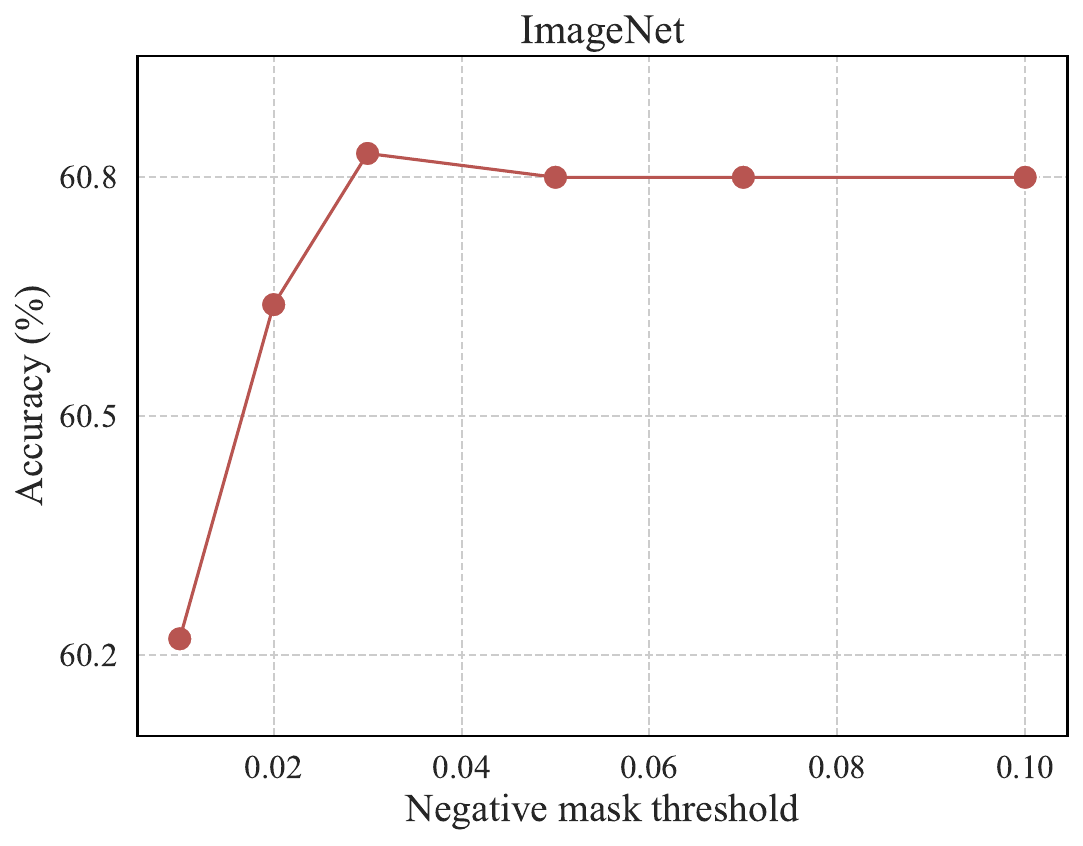}
  \caption{
 Parameter studies on the \textit{Negative Mask Threshold} $p_l$ for the negative pseudo-labeling in \textit{Negative Cache}. The results are reported on ImageNet top-1 accuracy using only the \textit{Negative Cache} to produce an adapted prediction. The experiments are conducted with CLIP-ResNet50.
  }
  \label{fig:nm}
\end{figure}

\section{Parameter Studies on Thresholds}
This section provides more parameter studies on three thresholds defined in our work. Our experiments are conducted on the ImageNet validation set using the default settings.

\paragraph{The threshold for negative pseudo-labeling.} 
In Eq~4 of our manuscript, the threshold $p_l$ is used to select negative pseudo labels by applying the negative mask. We conduct parameter studies on $p_l$ and the results are illustrated in Figure~\ref{fig:nm}. The best performance is achieved when $p_l$ is set to 0.03 and subsequent increases in $p_l$ do not yield notable improvement or degradation in the results, indicating the stability of this parameter. It can be noticed that the performance deteriorates when $p_l$ is less than 0.03 because the confident classes with low probabilities should not be included in negative pseudo labels.

\paragraph{The threshold range for testing feature selection in the negative cache.} 
In Eq~5 of our manuscript, the thresholds [$\tau_l$, $\tau_h$] are used to check whether the testing feature will be considered to be included in \textit{Negative Cache} if the entropy of the prediction is in the specified interval.
Table \ref{tab:range} presents the results of an ablation study focusing on the impact of adjusting the threshold range for testing feature selection in the \textit{Negative Cache}. This investigation delves into the testing feature selection of uncertain samples using two distinct approaches: one involving values closer to the minimum range threshold ($\tau_l$) and the other to the maximum range threshold ($\tau_h$), achieved by reversing the second condition: $\mathrm{H}(f_{\rm test} \mathbf{W}_c^T) < \mathrm{H}(\mathbf{\tilde{q}}^{ent} \mathbf{W}c^T)$ or $\mathrm{H}(f_{\rm test} \mathbf{W}_c^T) > \mathrm{H}(\mathbf{\tilde{q}}^{ent} \mathbf{W}_c^T)$. 
By collecting the lowest entropy features within the range [0.2, 0.5], the highest result is attained 60.83\%. Opting values below 0.2 indicates the selection of confident samples for the \textit{Negative Cache}, resulting in a reduction in the confidence of CLIP's prediction. Furthermore, a shift by 0.1 from [0.2, 0.5] to [0.3, 0.6] in the thresholds leads to an inclusion of more noisy samples during the early collection phase of the negative cache, resulting in a 0.48\% decrease in performance. Selecting maximum entropy features with the same threshold range displays a slight decline in performance compared to the minimum entropy feature selection within the specified range. Hence, the most valuable uncertain samples fall within the [0.2, 0.5] range, whose entropy is closer to 0.2. The reported results exclusively utilize the \textit{Negative Cache} to generate an adapted prediction. 

\begin{table}[]
\renewcommand\arraystretch{1.25}
\begin{tabular}{cccccc}
\hline

\multicolumn{3}{c|}{Minimum entropy features} & \multicolumn{3}{c}{Maximum entropy features} \\ \hline
$\tau_l$ & $\tau_h$ & \multicolumn{1}{c|}{Accuracy} & $\tau_l$ & $\tau_h$ & Accuracy \\ \hline
0.0 & 1.0 & \multicolumn{1}{c|}{60.69} & 0.2 & 0.4 & 60.51 \\
0.0 & 0.2 & \multicolumn{1}{c|}{60.67} & \textbf{0.2} & \textbf{0.5} & \textbf{60.53} \\
0.0 & 0.3 & \multicolumn{1}{c|}{60.69} & 0.2 & 0.6 & 60.51 \\
0.1 & 0.3 & \multicolumn{1}{c|}{60.76} & 0.3 & 0.5 & 60.30 \\
0.1 & 0.4 & \multicolumn{1}{c|}{60.77} & 0.3 & 0.7 & 60.30 \\
0.2 & 0.4 & \multicolumn{1}{c|}{60.81} & 0.4 & 0.6 & 60.16 \\
\textbf{0.2} & \textbf{0.5} & \multicolumn{1}{c|}{\textbf{60.83}} & 0.4 & 0.7 & 60.16 \\
0.2 & 0.6 & \multicolumn{1}{c|}{60.81} & 0.5 & 0.7 & 60.34 \\
0.3 & 0.5 & \multicolumn{1}{c|}{60.34} & 0.5 & 0.8 & 60.34 \\
0.3 & 0.6 & \multicolumn{1}{c|}{60.35} & 0.6 & 0.8 & 60.35 \\ \hline
\end{tabular}
\caption{
Ablation study of the impact of varying \textit{Threshold Range} [$\tau_l$, $\tau_h$] for testing feature selection in the \textit{Negative Cache}. The study investigates the testing feature selection of the uncertain samples in two ways: choosing the minimum and maximum entropy features in the given range. The results are reported on ImageNet top-1 accuracy using only the \textit{Negative Cache} to produce an adapted prediction. The experiments are conducted with CLIP-ResNet50.}
\label{tab:range}
\end{table}

\paragraph{The residual and sharpness ratios.} 
The experiments in Table~\ref{tab:alphabeta} show that the optimal residual ratio is 2.0 for TDA (instead of 1.0 in Tip-Adapter), indicating a higher significance of adapted features compared with CLIP features in test-time adaptation. The optimal sharpness ratio for TDA is 5.0, which is close to the 5.5 in Tip-Adapter.

\renewcommand\arraystretch{1.5}
\begin{table}[!h]
\resizebox{\linewidth}{!}{
\begin{footnotesize}
\begin{tabular}{lcccccc}
\hline
\multicolumn{1}{l|}{Residual Ratio} & 0.5                       & 1.0                                & \textbf{2.0}                       & 3.0                       & 4.0                       & 5.0                       \\ \hline
\multicolumn{1}{l|}{TDA}                                         & \multicolumn{1}{l}{61.07} & \multicolumn{1}{l}{61.20}          & \multicolumn{1}{l}{\textbf{61.35}} & \multicolumn{1}{l}{61.29} & \multicolumn{1}{l}{60.90} & \multicolumn{1}{l}{60.63} \\ \hline \hline
\multicolumn{1}{l|}{Sharpness Ratio} & 0.5                       & 1.0                                & 3.0                       & \textbf{5.0}                       & 7.0                       & 9.0                       \\ \hline 
\multicolumn{1}{l|}{TDA}                                         & 60.98                     & 61.20                              & 61.29                     & \textbf{61.35}                     & 61.20                     & 61.19                     \\ \hline
\end{tabular}
\end{footnotesize}}
\vspace{-5pt}
\caption{Analysis on the residual and sharpness ratios of TDA.}
\label{tab:alphabeta}
\end{table}

\section{More Experimental Analysis}
\paragraph{Caches built for inference.}
The caches are built on the fly during inference, starting empty and progressively accumulating samples. At the start of the testing phase on ImageNet, where only 1\% of the data was used, we observed a slight accuracy drop of 0.06\%. We also noted that bypassing cache usage at the early testing phase leads to a marginal accuracy improvement of 0.1\%. We didn’t adopt this approach as it increases complexity by introducing an extra hyperparameter for determining when to use caches.

\paragraph{Class imbalance under high shot capacity.}
Our analysis with a 6-shot positive cache reveals minimal class imbalance (only 4 out of 1000 ImageNet classes have less than 6 samples) but identifies a significant cache accuracy drop from 90.3\% to 86.6\% when shot capacity increases from 3 to 6. Such accuracy drop happens because larger cache capacities tend to accumulate noise, thereby reducing the reliability of cached labels and negatively affecting the adapted predictions. Hence, the performance decline with larger caches is mainly due to noise accumulation rather than class imbalance.

\section{Broader Impact}
The broader impact of test-time adaptation of vision-language models lies in its potential to enhance real-world applicability, improve accessibility and inclusivity, address bias and fairness concerns, and advance research and development. By allowing models to adapt to new, unseen data during inference, these models can be more versatile and adaptable, benefiting various domains such as healthcare and assistive technologies. Test-time adaptation also offers opportunities to mitigate biases, personalize user experiences, and push the boundaries of what vision-language models can achieve. However, ethical considerations must be taken into account to ensure responsible development and deployment, ensuring transparency, fairness, and accountability in the adaptation process.

\end{document}